\pdfoutput=1
\documentclass[11pt]{article}
\usepackage[preprint]{acl}

\usepackage{graphicx} 
\usepackage{bm}
\usepackage{soul}
\usepackage{graphicx}
\usepackage{tikz}
\usepackage{forest}
\usetikzlibrary{trees,positioning,shapes,shadows,arrows.meta}
\usepackage{amsmath}
\usepackage{amssymb}
\usepackage{pifont}
\usepackage{amsfonts}
\usepackage{dsfont} 
\usepackage{multirow}
\usepackage{wrapfig}
\usepackage{enumitem}
\usepackage{colortbl} 
\usepackage{dsfont}
\usepackage{lipsum}   
\usepackage{transparent}

\usepackage{times}
\usepackage{latexsym}

\usepackage{comment}

\usepackage[T1]{fontenc}

\usepackage[utf8]{inputenc}

\usepackage{microtype}

\usepackage{inconsolata}

\usepackage{graphicx}

\newcommand{\blackcircled}[1]{%
  \tikz[baseline=(char.base)]{
    \node[shape=circle,draw,fill=black,inner sep=1.2pt] (char) {\textcolor{white}{\scriptsize #1}};
  }%
}

\title{A Survey of Uncertainty Estimation Methods on Large Language Models}

\author{\textbf{Zhiqiu Xia, Jinxuan Xu, Yuqian Zhang, Hang Liu}\\
Rutgers, The State University of New Jersey\\
\texttt{\{zhiqiu.xia, jinxuan.xu, yqz.zhang, hang.liu\}@rutgers.edu}
}

\begin{document}

\maketitle

\begin{abstract}
Large language models (LLMs) have demonstrated remarkable capabilities across various tasks. However, these models could offer biased, hallucinated, or non-factual responses camouflaged by their fluency and realistic appearance. Uncertainty estimation is the key method to address this challenge. While research efforts in uncertainty estimation are ramping up, there is a lack of \textit{comprehensive} and \textit{dedicated} surveys on LLM uncertainty estimation. This survey presents four major avenues of LLM uncertainty estimation. Furthermore, we perform extensive experimental evaluations across multiple methods and datasets. At last, we provide critical and promising future directions for LLM uncertainty estimation.
\end{abstract}

\section{Introduction}
\begin{figure}[t]
\centering
\begin{minipage}[t]{\linewidth}
  \centering
  \includegraphics[width=\textwidth]{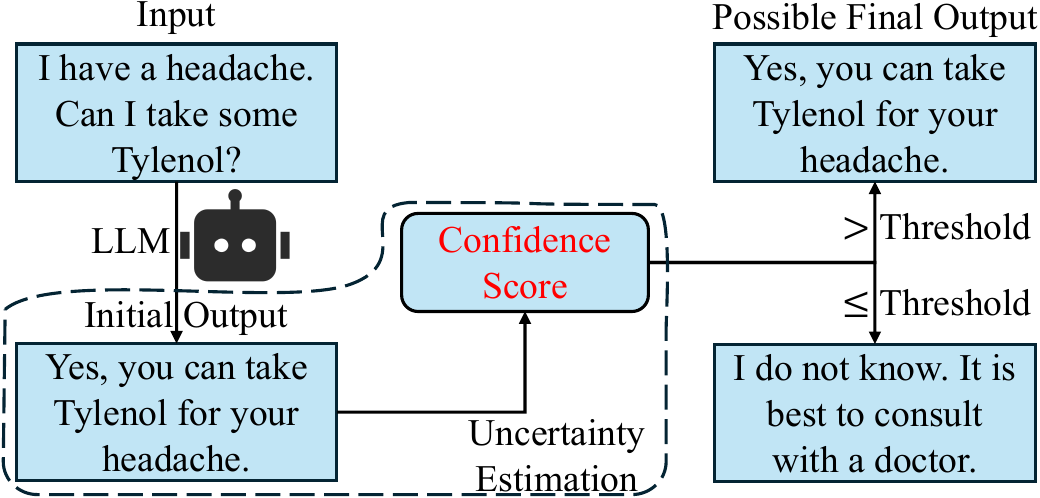}
  \vfill
    \caption{Illustration of uncertainty estimation.}
    \label{fig: ue}
\end{minipage}
\end{figure}

Large Language Models (LLMs) have emerged as state-of-the-art solutions for a wide range of problems, mainly due to their unparalleled ability to generate coherent and contextually appropriate responses to diverse user prompts~\cite{ouyang_advances_in_neural, zhao_a_survey_of}. However, with the increasing adoption of LLMs, concerns have grown regarding their tendency to produce biased, hallucinated, non-factual, and misaligned outputs~\cite{zhang_sirens_song_in, huang_a_survey_on}. These issues are further exacerbated by the fact that such flawed responses often appear highly fluent and convincingly realistic, making them difficult to detect.
A promising approach to addressing the challenge of misleading yet plausible responses is uncertainty estimation, which assigns an uncertainty or confidence score to the model's output. Figure~\ref{fig: ue} provides an overview of this process. First, the LLM generates an initial response based on the input. Next, a confidence score is computed for this response. The score is then evaluated against a predefined threshold to determine the final output. If the confidence score meets or exceeds the threshold, the initial response is accepted; otherwise, the model outputs "I do not know," thereby reducing the risk of providing incorrect but convincingly realistic information to users.

There is an urgent need for a comprehensive survey on LLM uncertainty estimation. Below, we highlight three of them: (i) Although uncertainty estimation has been extensively studied in traditional deep neural networks (DNNs)—with Bayesian and ensemble methods being notable examples~\cite{gawlikowski2023survey})—these techniques are not easily transferable to LLMs, due to the large number of parameters in LLMs.
(ii) LLMs significantly transform society, creating a strong demand for a thorough study of uncertainty estimation tailored to LLMs. While numerous uncertainty estimation methods exist, recent benchmark studies~\cite{fadeeva2023_lm_polygraph, vashurin2025benchmarking_uncertainty} have focused on their empirical evaluation rather than in-depth methodological discussions. Consequently, a survey of recent advances in LLM uncertainty estimation that synthesizes recent progress is crucial, offering a solid foundation for future development in the field.
(iii) Besides surveys focusing on natural language processing~\cite{fomichev2020unsupervised_quality, baan2023uncertainty_in}, there are three existing surveys concentrating on LLM uncertainty estimation. However, each has notable limitations. Specifically, \citet{huang_a_survey_of} dedicates a substantial portion of its content to traditional DNN uncertainty estimation rather than focusing on LLMs. \citet{geng_a_survey_of} shifts its attention to uncertainty calibration and the applications of LLM uncertainty estimation, rather than providing a deep exploration of the core techniques. Similarly, \citet{shorinwa_a_survey_on} devotes much of its content to benchmarks and applications while lacking a complete view of the uncertainty estimation methods on LLMs.

This work focuses on studying the uncertainty estimation methods within the context of LLMs, introducing a new taxonomy from the perspective of LLMs. We center our scope around techniques applicable during the inference stage. 
We focus on the uncertainty estimation methods, excluding confidence calibration methods~\cite{zhou2023navigating_the, detommaso2024multicalibration} from our scope.
Besides, we emphasize the methods that do not require additional data \cite{ren_out_of_distribution, kumar_conformal_prediction_with, tonolini_bayesian_prompt_emsembles} or model modifications \cite{huang_large_language_models, liu_uncertainty_estimation_and}, ensuring the broad applicability of this survey. 
Furthermore, this survey conducts a thorough evaluation of representative uncertainty estimation approaches across various datasets and domains. Built on the insights from our evaluations, we postulate two interesting future directions for LLM uncertainty estimation.

\section{Uncertainty Sources in LLM} \label{sec: uncertainty_source}

There are two primary sources of uncertainty: aleatoric and epistemic uncertainties~\cite{kendall_advances_in_neural, hullermeier_aleatoric_and_epistemic}. In the context of LLMs~\cite{gao_spuq, ahdritz_distinguishing_the_knowable, hou_decomposing_uncertainty_for}, these sources manifest in the following ways:

\begin{itemize}
    \item \textbf{Aleatoric uncertainty} refers to the uncertainty inherent in the data. For LLMs, this arises from ambiguous or incomplete information and inherent properties of natural language itself. Examples include vague or contextually dependent prompts, as well as linguistic phenomena where multiple valid interpretations or responses naturally coexist.
    \vspace{-.1in}
    \item \textbf{Epistemic uncertainty} reflects the model's lack of knowledge or understanding. In LLMs, this occurs when the model encounters unfamiliar concepts or data that are underrepresented in its training set. This type of uncertainty can potentially be reduced by improving the training datasets and models. 
\end{itemize}

\section{Uncertainty Estimation in LLMs} \label{sec: uncertainty estimation in LLMs}
\subsection{Problem Definition and Overview} \label{sec: prob_def}
\textbf{Token generation in LLMs.} LLMs output responses in an auto-regressive manner, predicting the probability distribution of the next token given the prompt and the previously generated tokens. We denote the model as $\bm{f}$, the prompt as $\bm{x}$, and the generated response (or the answer) as $\bm{r}$, which consists of $N$ tokens, denoted as $\{z_1, z_2, z_3, \cdots, z_N\}$. The tokens can be either words, subwords, or characters from a predefined vocabulary $\bm{Z}$. At each step of token generation, the model computes the conditional probability distribution over the vocabulary for the next token, based on the prompt $\bm{x}$ and all previously generated tokens $\bm{r_{<i}} = \{z_1, z_2, \cdots, z_{i-1}\}$. The probability distribution for the $i$-th token is given by $\bm{p_i} = \text{Softmax}(\bm{f}(\bm{x}, \bm{r_{<i}}))$. Here, $\bm{p_i}$ is a vector of length $|\bm{Z}|$, with each entry representing the probability of a specific token in $\bm{Z}$ being chosen as the next token.
It allows strategies such as sampling or beam search to choose from these token candidates according to their probabilities. Such an auto-regressive process ends when \# of generated tokens reaches a preset number or LLM generates the end-of-sequence (EOS) token. 

\begin{figure}[h]
\centering
\begin{minipage}[t]{\linewidth}
  \centering
  \includegraphics[width=\textwidth]{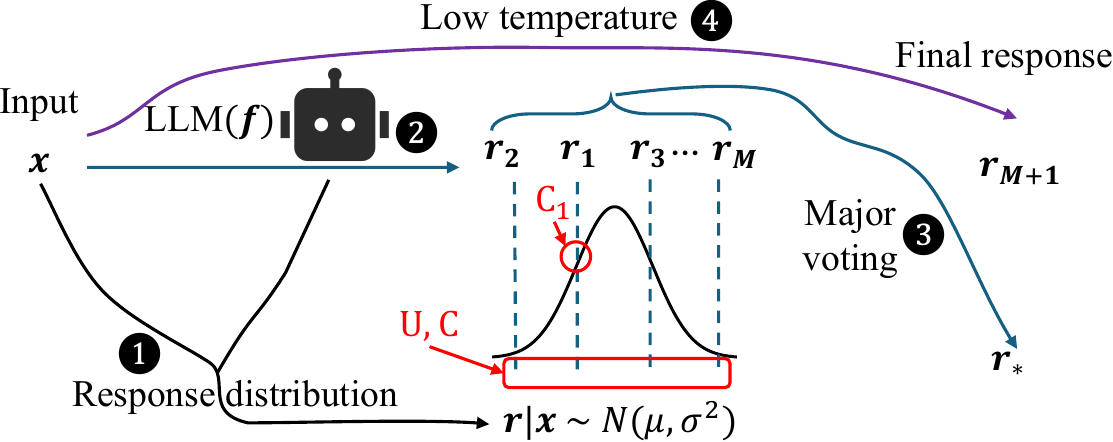}
  \vfill
    \caption{Illustration of uncertainty versus confidence.}
    \label{fig: gaussian}
\end{minipage}
\end{figure}

\begin{figure*}
    \centering
    
\tikzset{
    basic/.style  = {draw, text width=3cm, align=center, font=\sffamily, rectangle, font=\small},
    root/.style   = {basic, rounded corners=2pt, thin, align=center, text width=6cm, rotate=90},
    tnode/.style = {basic, thin, align=left, fill=cyan!10, text width=10.8cm, align=left},
    xnode/.style = {basic, thin, rounded corners=2pt, align=center, fill=green!20,text width=2.7cm,},
    wnode/.style = {basic, thin, align=left, fill=cyan!10, text width=2.7cm, align=left},
    onode/.style = {basic, thin, align=left, fill=cyan!10, text width=8.1cm, align=left},
}
\begin{forest} for tree={
    grow=east,
    growth parent anchor=west,
    parent anchor=east,
    child anchor=west,
    edge path={\noexpand\path[\forestoption{edge},->, >={latex}] 
         (!u.parent anchor) -- +(10pt,0pt) |-  (.child anchor) 
         \forestoption{edge label};}
}
[Uncertainty Estimation Methods on LLMs, root, l sep=6mm, parent anchor=south
    [Semantic Clustering\\Methods (\S~\ref{sec: se}), xnode,  l sep=6mm,
        [Implicit clustering: \cite{duan_shifting_attention_to} \cite{fadeeva2024fact_checking} \cite{lin_generating_with_confidence} \cite{nikitin_kernel_language_entropy}, tnode]
        [Explicit clustering: \cite{kuhn_semantic_uncertainty_linguistic} \cite{farquhar_detecting_hallucinations_in} , tnode] ] 
    [Consistency-based\\Methods (\S~\ref{sec: consistency}), xnode, l sep=6mm,
        [Similarity-based: \cite{huang_look_before_you} \cite{manakul_selfcheckgpt} \cite{gao_spuq} \cite{wang2024on_subjective} 
        \cite{chen_quantifying_uncertainty_in} \cite{zhang_luq_long_text}
        \cite{tanneru_quantifying_uncertainty_in}  
        \cite{lin_generating_with_confidence},
        tnode]
        [Agreement-based: \cite{cole_selectively_answering_ambiguous} \cite{lyu_calibrating_large_language} \cite{hou_decomposing_uncertainty_for}, tnode]]
    [Latent Information\\Methods (\S~\ref{sec: latent information}), xnode,  l sep=6mm,
        [Hidden states-based: \cite{chen_inside_llms_internal} \cite{Sriramanan2024llm_check} , tnode]
        [Probability distribution-based: \cite{manakul_selfcheckgpt} \cite{zhang2023enhancing_uncertainty} \cite{jiang_calibrating_language_models} \cite{ahdritz_distinguishing_the_knowable}, tnode]
        [Predicted probability-based:
            \cite{jiang_how_can_we}
            \cite{manakul_selfcheckgpt}
            \cite{kadavath_language_models_mostly} 
            \cite{protillo_strength_in_numbers}
            \cite{ling_uncertainty_quantification_for}
            \cite{malinin_uncertainty_estimation_in} 
            \cite{bakman_mars_meaning_aware}
            , tnode
        ]] 
    [Verbalizing\\Methods (\S~\ref{sec: verbalizing}), xnode,  l sep=6mm,
        [Heuristic: \cite{tian_just_ask_for} \cite{tanneru_quantifying_uncertainty_in} \cite{xiong_can_llms_express}, tnode] ]
]
\end{forest}
    \caption{Taxonomy of uncertainty estimation methods on LLMs.}
    \label{fig:lit_surv}
\end{figure*}
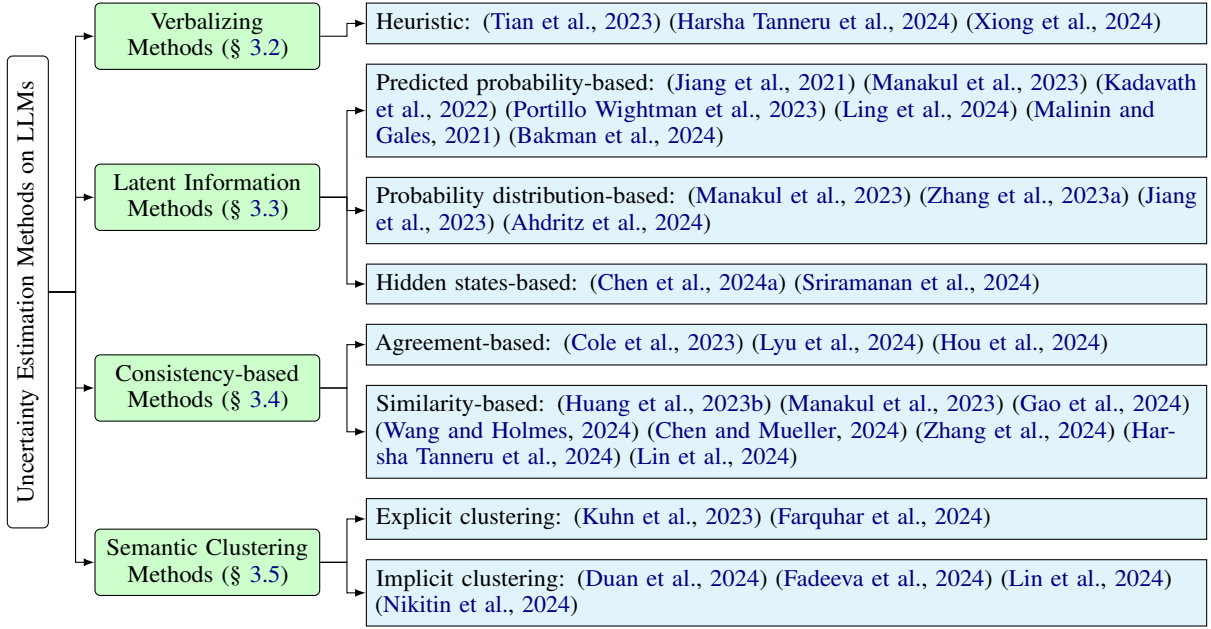

It is important to note that uncertainty is the innate nature of LLMs, regardless of whether we estimate it. Now, we provide an intuitive understanding of uncertainty and how to estimate it.

\vspace{.1in}
\noindent \textbf{How to estimate uncertainty and confidence? }
Following the conceptualization by~\citet{lin_generating_with_confidence}, we illustrate the process as shown in Figure~\ref{fig: gaussian}.
For each input $\bm{x}$, an LLM model has an underlying response distribution for it (\blackcircled{1}). For ease of illustration, we assume the distribution is a normal distribution $N(\mu, \sigma^2)$. Uncertainty estimation is to estimate the underlying variance $\sigma^2$. For example, the sample variance of $M$ different responses $\bm{r_1}, \cdots, \bm{r_M}$ (\blackcircled{2}) can be an estimator for the variance, which indicates the variations of responses (\textcolor{red}{$U$} in Figure~\ref{fig: gaussian}).

There are generally two types of confidence, i.e., overall confidence $C$ and the confidence $C_i$ associated with each response candidate $\bm{r_i}$. The overall confidence $C$ is complementary to $U$, i.e., the precision $1/\sigma^2$ of the distribution is a confidence $C$ to the input. The associated confidence is related to $\bm{x}$ and the tokens in a specific response $\bm{r_i}$. 
To provide the final response to answer the input $\bm{x}$ given sampled responses, some literature resort to majority voting to select the most-voted response $\bm{r}_{*}$ (\blackcircled{3})~\cite{wang_self_consistency_improves}, while others choose to generate one extra response  $\bm{r_{M+1}}$ with low-temperature settings (\blackcircled{4})~\cite{farquhar_detecting_hallucinations_in}.

\vspace{.1in}
\noindent
\textbf{Survey papers overview.}
Figure~\ref{fig:lit_surv} categorizes all the uncertainty estimation papers for LLM into four classes: verbalizing methods, latent information methods, consistency-based methods, and semantic clustering methods. We review each of them through Sections~\ref{sec: verbalizing} -~\ref{sec: se}.

\subsection{Verbalizing Methods} \label{sec: verbalizing}


\begin{figure}[htbp]
\centering
\begin{minipage}[t]{\linewidth}
  \centering
  \includegraphics[width=\textwidth]{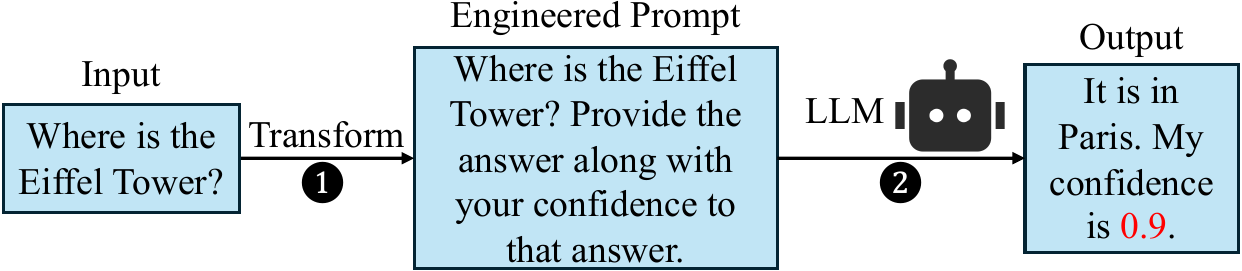}
  \vfill
    \caption{Illustration on verbalizing methods.}
    \label{fig: verbalizing_draw}
\end{minipage}
\end{figure}

Figure~\ref{fig: verbalizing_draw} demonstrates the main workflow of verbalizing methods. Firstly, the input is transformed into an engineered prompt that explicitly asks the model to provide both an answer and its confidence level (\blackcircled{1}). Secondly, the LLM processes this prompt and generates an output that includes the answer and a verbalized confidence score (\blackcircled{2}), representing its self-assessed certainty about the correctness of its response. 

\citet{lin_teaching_models_to} pioneers this cohort of efforts. As the capabilities of LLMs continue to develop, they can provide reasonable confidence under proper guidance, even without fine-tuning. 
Subsequently, \citet{tian_just_ask_for} proposes three verbalizing variants: (i) Generate multiple response candidates with confidence scores and select the highest-rated one as the final response, (ii) derive the response and confidence through two rounds of prompt-and-answer interactions, and (iii) use words instead of numerical values to indicate the confidence. 
Recently, \citet{tanneru_quantifying_uncertainty_in} introduces two methods inspired by Chain-of-Thought (CoT) prompting. The first method requests the LLM to assign an importance score to each word in the input, while the second one prompts the LLM to provide confidence for each reasoning step in the response. Finally, LLM will offer a final confidence score for the overall response. 
Beyond that, 
\citet{xiong_can_llms_express} presents a systematic framework for verbalizing methods with three parts: prompting, sampling, and aggregation. It employs specific confidence-eliciting prompts and generates diverse response samples containing confidence scores. After that, the final confidence score is derived through inter-sample agreement or response ranking information.

While verbalizing methods offer intuitive and straightforward uncertainty estimation, they face significant limitations. \citet{kadavath_language_models_mostly} shows that LLMs tend to be over-confident in their answers as the reinforcement learning from human feedback (RLHF) nature pushes LLMs to do so. 

\subsection{Latent Information Methods} \label{sec: latent information}


\begin{figure}[htbp]
\centering
\begin{minipage}[t]{\linewidth}
  \centering
  \includegraphics[width=\textwidth]{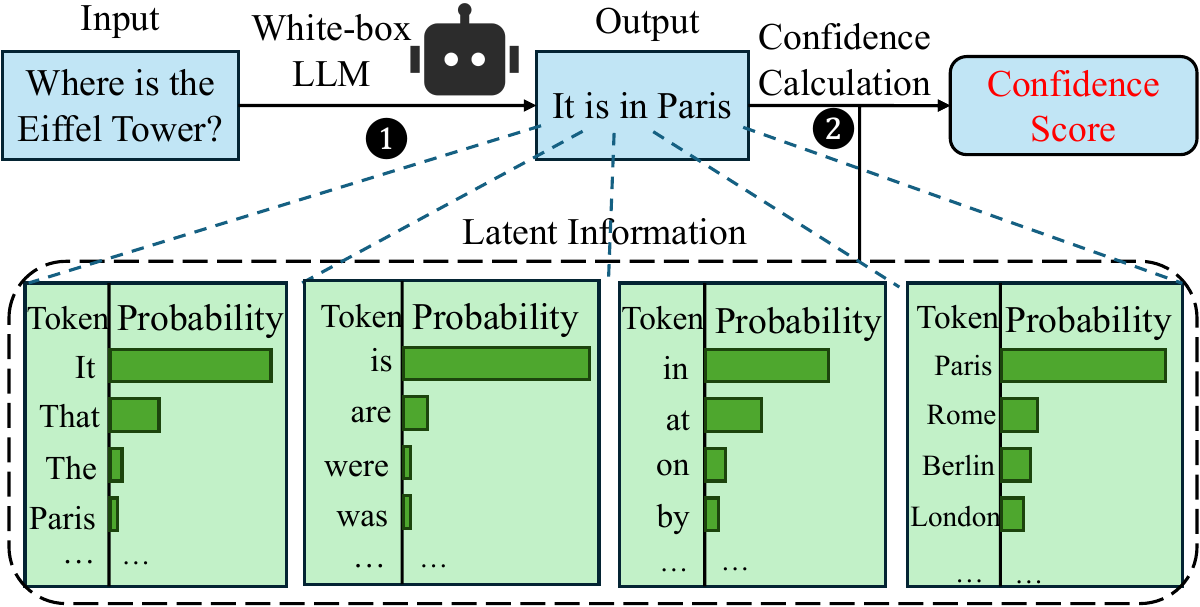}
  \vfill
    \caption{Illustration on latent information methods.}
    \label{fig: latent_draw}
\end{minipage}
\end{figure}

Figure~\ref{fig: latent_draw} illustrates the concept of latent information methods. First of all, the LLM is prompted to provide an output to the input (\blackcircled{1}). Of note, latent information methods require a white-box LLM, which offers latent information in the output, such as the full probability distribution over each generated token. Subsequently, this method leverages the generated information to estimate the uncertainty/confidence score via specific metrics or measures (\blackcircled{2}). We refer the readers to Section~\ref{app:formula_latent} for the formula of different latent information methods.

\citet{jiang_how_can_we} directly uses the predicted probability of the response tokens to measure the confidence score. \citet{manakul_selfcheckgpt} proposes to use the negative log-likelihood of the response tokens, either average or maximum across tokens, to serve as an uncertainty measure. The averaged negative log-likelihood across tokens is also known as perplexity \cite{ren_out_of_distribution}.
In contrast, \citet{kadavath_language_models_mostly} proposes a method that prompts the model to evaluate its answers by answering true or false, using the latent probability associated with ``True'' as the confidence score. 

The analysis of token probabilities can be extended beyond a single response for more robust uncertainty estimation.  
\citet{protillo_strength_in_numbers} proposes to average the predicted probabilities across multiple responses. 
\citet{ling_uncertainty_quantification_for} picks the key token from the responses and aggregates them into a distribution, and the uncertainty is from the entropy of the distribution.
\citet{kadavath_language_models_mostly} considers all the tokens in the responses, calculates the probability for each response using token probabilities, and measures uncertainty through the entropy of the response distribution, called predictive entropy. 
However, varying response lengths can introduce undesirable noise to the estimation. To address this limitation, \citet{malinin_uncertainty_estimation_in} proposes the length-normalized entropy, incorporating the response length based on predictive entropy. 
Furthermore, \citet{bakman_mars_meaning_aware} proposes to replace the length normalization by assigning a weight to each token with a BERT model to consider both the sequence length and the semantic contribution of tokens.

While the methods above only require access to the probability value of the response tokens, the following papers would require access to the complete probability distributions: 
\citet{manakul_selfcheckgpt} computes the entropy of the probability distribution for each generated token, using either the mean or maximum entropy as the uncertainty.
\citet{zhang2023enhancing_uncertainty} proposes an uncertainty metric that combines the sum of negative log probabilities and entropy, while also considering token importance, preceding context, and token properties such as entity type and token frequency.
For multiple-choice questions, \citet{jiang_calibrating_language_models} presents a specialized methodology. It computes probability distributions over potential options for each response sample and aggregates these distributions to form an ensemble probability distribution for uncertainty estimation.
\citet{ahdritz_distinguishing_the_knowable} introduces a heuristic two-stage method. Initially, the LLM is prompted to generate multiple next-token candidates. Subsequently, through a "repeated prompt" mechanism, the model produces the next token. The final uncertainty score is then computed from the probability distribution of these next tokens.

Beyond the methods using the probability distributions of tokens in the response, some researchers utilize the hidden states of LLMs. 
\citet{chen_inside_llms_internal} proposes to use the embeddings in the middle layer of LLMs to construct a covariance matrix for responses, which captures the correlation relationships among them. By manipulating the eigenvalues of the covariance matrix, the degree of divergence among responses can be estimated and considered an uncertainty measure. Different from this approach, \citet{Sriramanan2024llm_check} constructs a covariance matrix within a single response, where each row corresponds to a token in the response. The log-determinant of this covariance matrix is then calculated as an uncertainty metric. Additionally, it proposes another method based on the internal components of the attention mechanism. Specifically, it calculates the sum of the log-determinants of each self-attention head’s kernel similarity map, which serves as an alternative measure of uncertainty.

\subsection{Consistency-based Methods} \label{sec: consistency}


\begin{figure}[htbp]
\centering
\begin{minipage}[t]{\linewidth}
  \centering
  \includegraphics[width=\textwidth]{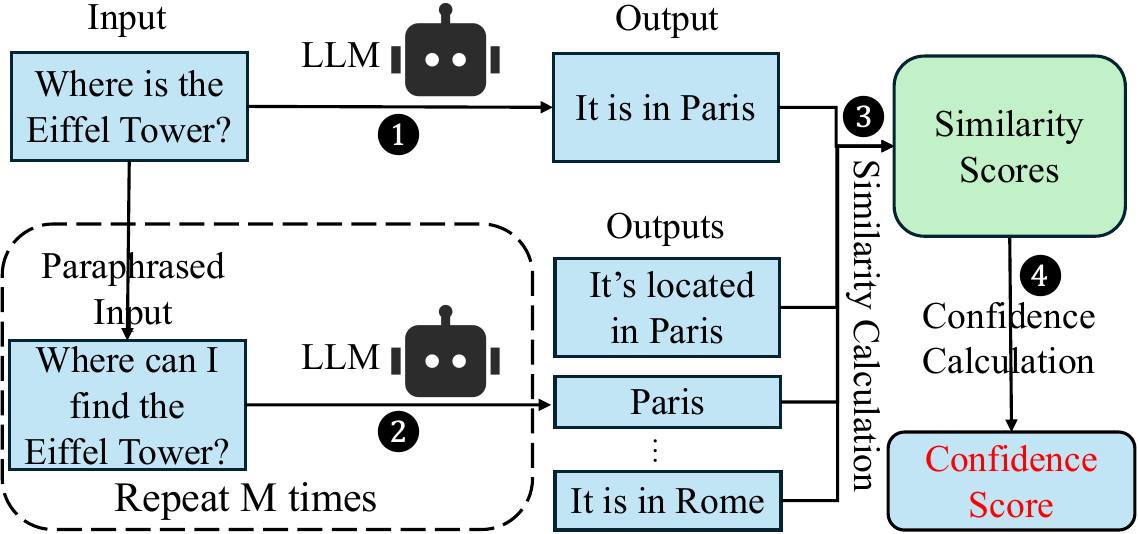}
  \vfill
    \caption{Illustration of consistency-based methods.}
    \label{fig: consistency_draw}
\end{minipage}
\end{figure}

Figure~\ref{fig: consistency_draw} illustrates the workflow of consistency-based methods. First, LLM gives an output to the original input (\blackcircled{1}). Second, the input is paraphrased to maintain the same meaning as the original one but has different contents, where LLM is prompted to answer this changed input. Such process is repeated $M$ times to generate various sampled outputs (\blackcircled{2}). Third, the similarities between the original output and each sampled output are computed (\blackcircled{3}). Finally, the confidence score is calculated based on derived similarities (\blackcircled{4}). 
We refer the readers to Section~\ref{app:formula_consis} for detailed mathematical definitions of this consistency-based method.

The fundamental principle of consistency-based methods is that response consistency typically correlates with confidence levels, a.k.a. high response variability suggests higher uncertainty, while consistent responses indicate greater confidence. 

\citet{cole_selectively_answering_ambiguous} introduces sampling diversity and sampling repetition. Sampling diversity quantifies the ratio of unique answers to the total number of samples, while sampling repetition measures the proportion of samples that align with the most frequent answer. Extending this framework, \citet{lyu_calibrating_large_language} enhances the sampling repetition metric by incorporating the most frequent and second-most frequent responses in its analysis.
\citet{hou_decomposing_uncertainty_for} presents a more nuanced approach by introducing clarification-based uncertainty estimation. It first generates multiple clarifications for the input and then produces responses based on these clarified inputs. 
The estimated uncertainty combines two parts: one from answer frequency distribution and the other from input clarification variance.

While the methods primarily focus on analyzing answer agreement patterns to estimate uncertainty, more methods emphasize evaluating the similarities among responses (\blackcircled{3}).
For domain-specific tasks, targeted metrics like BLEU \cite{papineni_bleu_a_method} and CodeBLEU \cite{ren_codebleu_a_method} have been successfully applied to machine translation and code generation tasks, respectively \cite{huang_look_before_you}.
In general question-answering scenarios, token-level similarity metrics such as BERTScore \cite{zhang_bertscore_evaluating_text} and RougeL \cite{lin_rouge_a_package} have been widely adopted \cite{huang_look_before_you, manakul_selfcheckgpt, gao_spuq}.
Moving beyond token-level comparisons, more sophisticated approaches that capture semantic relationships have emerged, including SentenceBERT and NLI-based methods \cite{wang2024on_subjective,gao_spuq, chen_quantifying_uncertainty_in, zhang_luq_long_text}.
SentenceBERT computes the cosine similarity between two sentences using embeddings generated by the Sentence Transformer model.
The NLI-based method leverages natural language inference (NLI) classifiers to categorize sentence relationships as entailment, neutral, or contradiction, regarding the probability the NLI classifier assigns to the ``entailment'' class as the similarity score.
Moreover, \citet{tanneru_quantifying_uncertainty_in} proposes token importance uncertainty and CoT uncertainty. The former quantifies uncertainty through token agreement and token rank metrics, while the latter evaluates inter-step relationships using NLI classification techniques.

The generation of diverse LLM outputs in step \blackcircled{2} represents another critical avenue for enhancing consistency-based methods. \citet{tanneru_quantifying_uncertainty_in} presents two fundamental approaches: sample probing, which employs semantically equivalent prompts, and model probing, which manipulates temperature settings to introduce output stochasticity. 
\citet{chen_quantifying_uncertainty_in} introduces a method that modifies CoT steps specifically for prompts employing CoT techniques. Additional approaches have been proposed by \citet{gao_spuq}, including the strategic insertion of dummy tokens (such as newline characters and tab spaces) and modifications to system messages within prompts.

While most methods estimate confidence by averaging similarities among responses in step (\blackcircled{4}), 
\citet{lin_generating_with_confidence} proposes using the number of semantic sets as a measure of uncertainty. These semantic sets are defined as ``semantic equivalent'' subsets, which are grouped from all responses using an NLI classifier. 

Notably, consistency-based methods are computationally expensive due to the need for multiple inferences. While some approaches rely on auxiliary modules, such as an NLI classifier, their computational cost is considerably lower than that of performing an LLM inference. As a result, the computational overhead is primarily dominated by the number of inferences required.

\subsection{Semantic Clustering Methods} \label{sec: se}


\begin{figure}[htbp]
\centering
\begin{minipage}[t]{\linewidth}
  \centering
  \includegraphics[width=\textwidth]{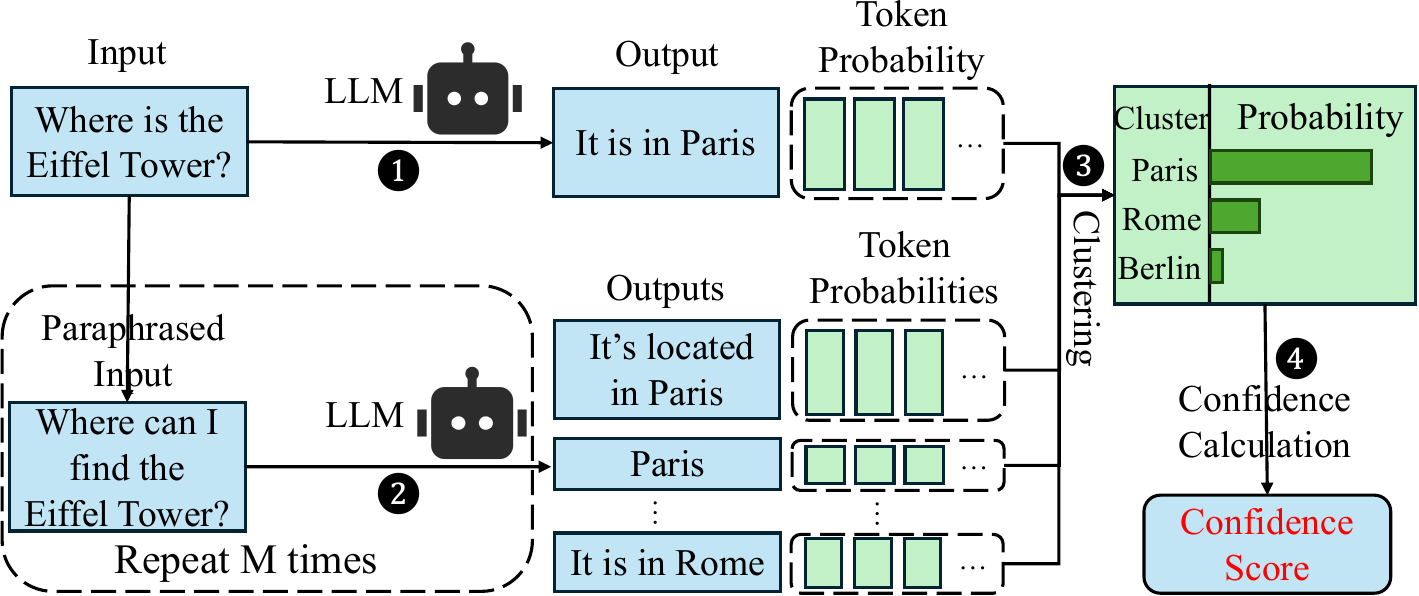}
  \vfill
    \caption{Illustration on semantic clustering methods.}
    \label{fig: semantic_draw}
\end{minipage}
\end{figure}

Figure~\ref{fig: semantic_draw} depicts the workflow of semantic clustering methods, which leverages both the latent information and the semantic relationships among responses to offer a more comprehensive estimation of the uncertainty.
The first two steps are similar to the consistency-based methods, where the LLM generates responses to the original input and its paraphrased versions (\blackcircled{1}-\blackcircled{2}). Next, instead of calculating the similarities, the sampled outputs are partitioned into clusters with a new probability for each cluster (\blackcircled{3}). Finally, the probability distribution over these clusters calculates a confidence score (\blackcircled{4}). The motivation for semantic clustering methods stems from the limitations of consistency-based approaches, which can only deliver a perfect uncertainty score when the responses use identical wording. However, in reality, responses may convey the same meaning through varied expressions. 
Therefore, semantic clustering of the responses is proposed to deal with the limitations. We refer the readers to Section~\ref{app:formula_semantic} for the formula of different semantic clustering methods.

\citet{kuhn_semantic_uncertainty_linguistic} introduces semantic entropy for uncertainty estimation. The method comprises three phases: generation (\blackcircled{1}-\blackcircled{2}), clustering (\blackcircled{3}), and entropy estimation (\blackcircled{4}). 
In step \blackcircled{3}, a bi-directional entailment algorithm is employed to cluster semantically equivalent responses. It assesses the entailment relationship between each pair of responses, considering them to express the same meaning if they mutually entail each other. The entailment relationship can be determined with the help of an NLI classifier or by simply requesting a general-purpose LLM. 
The uncertainty is the entropy calculated from the cluster probabilities in step \blackcircled{4}. In case there is no access to the token probability, \citet{farquhar_detecting_hallucinations_in} introduces discrete semantic entropy, extending the work of \citet{kuhn_semantic_uncertainty_linguistic}, which leverages the frequency of each cluster to calculate an entropy as the uncertainty. 

While the above method clusters the responses explicitly, some methods propose implicit clustering.
\citet{duan_shifting_attention_to} introduces sentence relevance scores between each response pair, which is more effective over long sentences than the bi-directional entail algorithm in the work of \cite{kuhn_semantic_uncertainty_linguistic}. \citet{fadeeva2024fact_checking} proposes to further alleviate the impact of claim-type uncertainty by grouping words into several claim types. Besides, it only requires one-time inference by selecting the top-k choices of words using the latent information to generate different responses.

In contrast to these latent-information-dependent techniques, some semantic clustering methods operate without requiring access to such information.
\citet{lin_generating_with_confidence} proposes to treat generated responses as nodes and obtain the degree matrix and the graph Laplacian matrix from the pairwise similarities of the responses. Correspondingly, this method defines several uncertainty and confidence measures from the matrices.
\citet{nikitin_kernel_language_entropy} further considers the distances between the clusters. The method encodes similarities among responses via positive semidefinite unit trace kernels. It offers a more fine-grained uncertainty measure using the von Neumann entropy of these kernels.

\section{Evaluation} \label{sec: evaluation}
\subsection{Metrics}

\begin{figure*}[htbp]
\centering
\begin{minipage}[t]{0.49\linewidth}
  \centering
  \includegraphics[width=\textwidth]{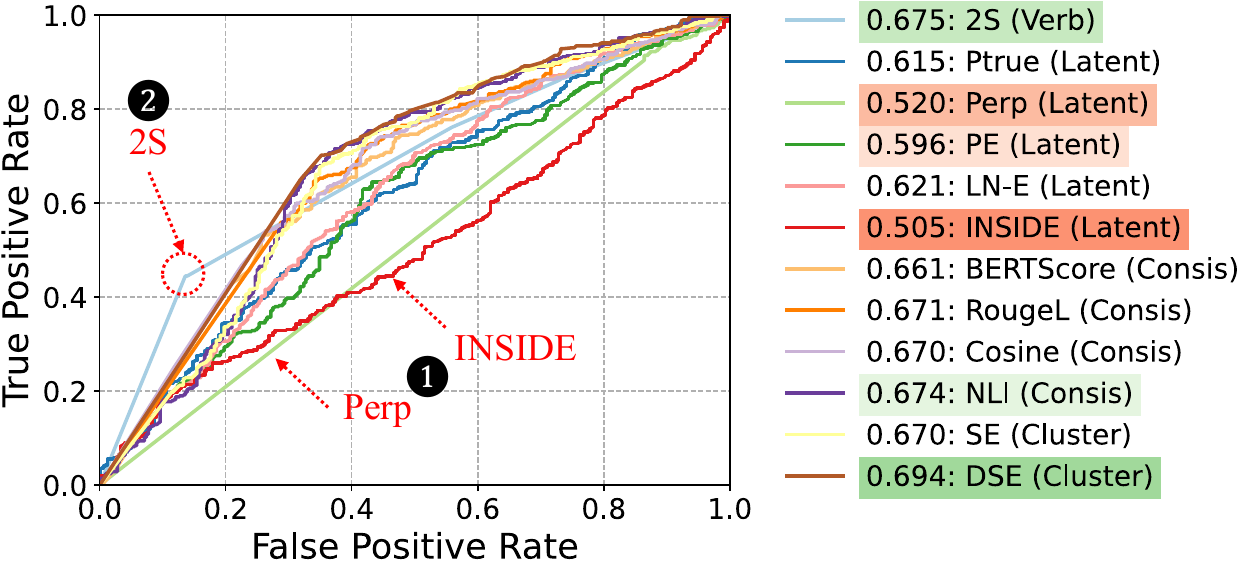}
  \vfill
\end{minipage}\hspace{0.01\linewidth}
\begin{minipage}[t]{0.49\linewidth}
  \centering
  \includegraphics[width=\textwidth]{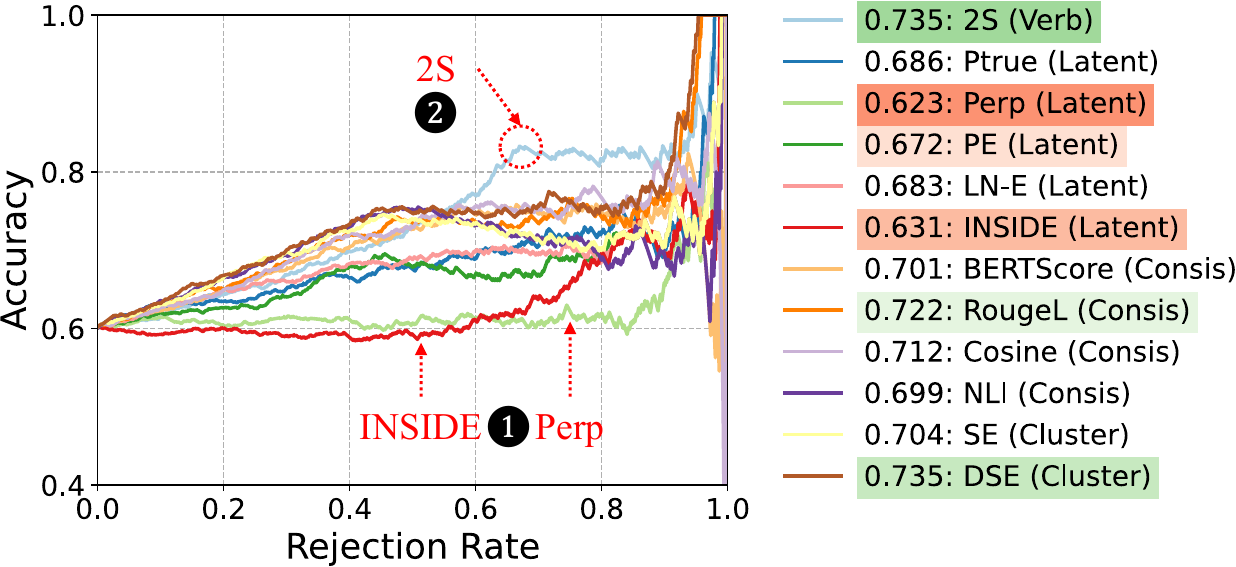}
  \vfill
\end{minipage}
\vspace{-.3in} 
\caption{\textbf{TruthfulQA}: ROC (left), ARC (right) curves, and AUROC and AUARC.}
\vspace{-.1in}
\label{fig: truthful_qa}

\vspace{1.5em}

\begin{minipage}[t]{0.49\linewidth}
  \centering
  \includegraphics[width=\textwidth]{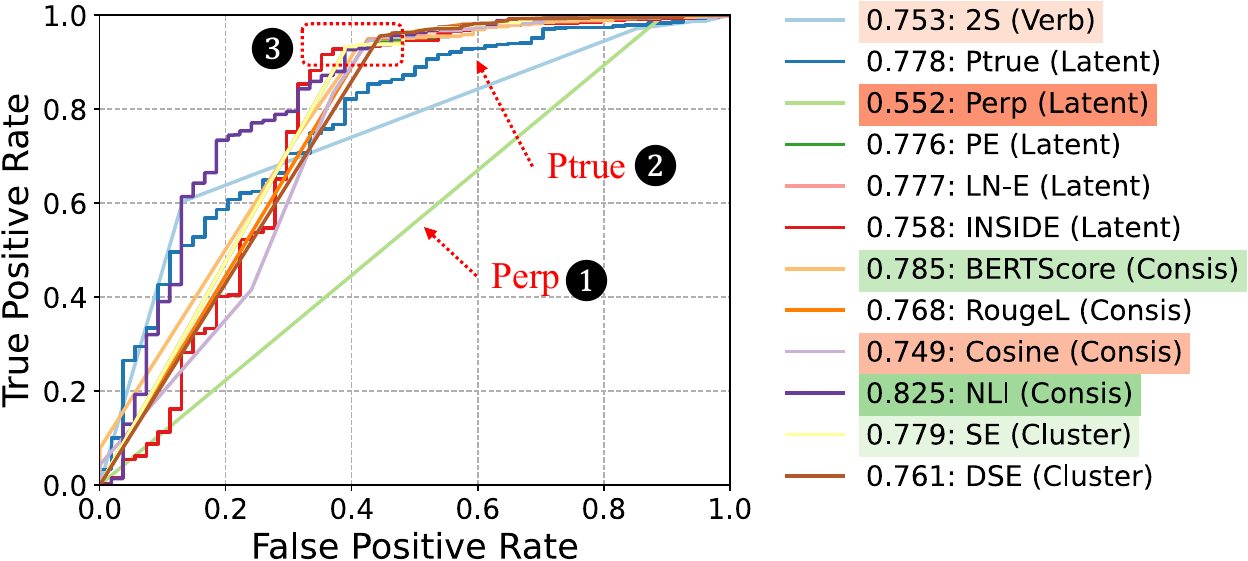}
  \vfill
\end{minipage}\hspace{0.01\linewidth}
\begin{minipage}[t]{0.49\linewidth}
  \centering
  \includegraphics[width=\textwidth]{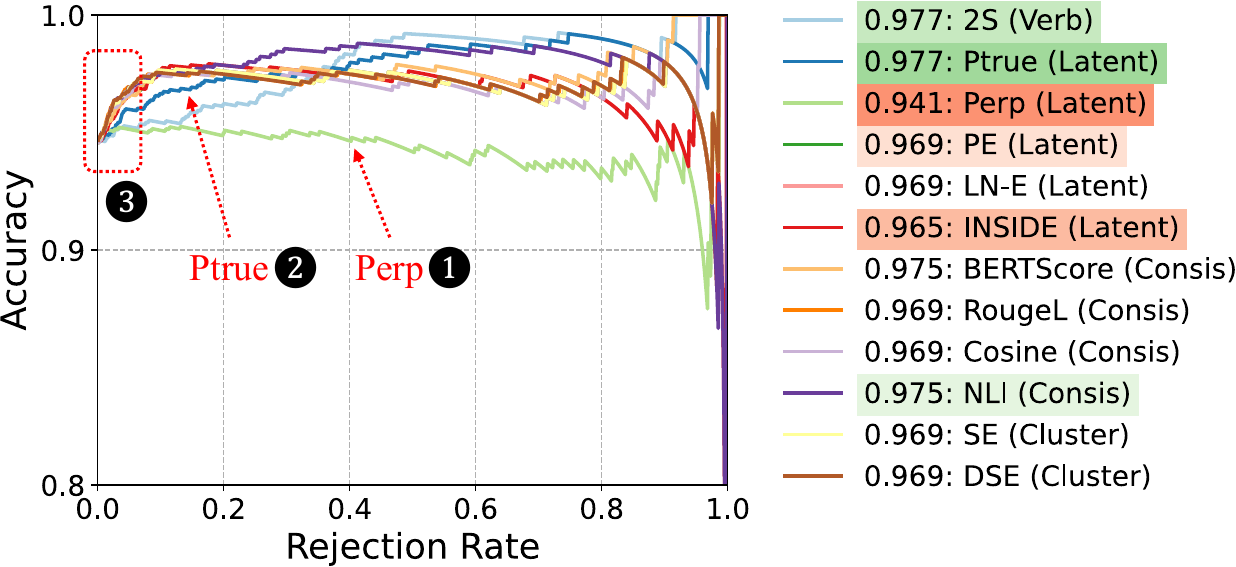}
  \vfill
\end{minipage}
\vspace{-.3in} 
\caption{\textbf{SciQ}: ROC (left), ARC (right) curves, and AUROC and AUARC.}
\vspace{-.1in}
\label{fig: sciq}
\end{figure*}

We use two primary metrics to evaluate the uncertainty estimation: \textbf{AUROC} (Area Under the Receiver Operating Characteristics curve) \cite{bradley_the_use_of} and \textbf{AUARC} (Area Under the Accuracy-Rejection Curve) \cite{nadeem_accuracy_rejection_curves}. Both metrics range from 0 to 1, with higher scores reflecting better uncertainty estimation methods. {\S~\ref{app:detail_au}} contains more details about AUROC and AUARC. Of note, while some research~\cite{huang2024uncertainty_in, chen2024reconfidencing_llms} propose novel metrics for uncertainty estimation, they focus more on the calibration ability. In alignment with established practices in the field~\cite{kuhn_semantic_uncertainty_linguistic, farquhar_detecting_hallucinations_in}, our evaluation primarily emphasizes a method's proficiency in discriminating between correct and incorrect answers based on the estimated uncertainty.

\subsection{Evaluated Methods} \label{sec: evaluated_methods}

We select several representative methods from each method category as follows:
\begin{itemize}
    \item \textit{Verbalizing methods (Verb):} We evaluate the  \textbf{2S}~\cite{tian_just_ask_for} method that asks for confidence in a second-round dialogue.
    \item \textit{Latent information methods (Latent):} We select the self-evaluation method (\textbf{Ptrue}) \cite{kadavath_language_models_mostly}, perplexity (\textbf{Perp}), predictive entropy (\textbf{PE}), length-normalized entropy (\textbf{LN-E}), and the method leveraging hidden states of LLMs (\textbf{INSIDE}) \cite{chen_inside_llms_internal}.
    \item \textit{Consistency-based methods (Consis):} We adopt four similarity measures: \textbf{BERTScore}, \textbf{RoughL}, cosine similarity from BERT embeddings (\textbf{Cosine}), and the ``entailment'' probability from an NLI classifier (\textbf{NLI}). The confidence score is averaged from similarities.
    \item \textit{Semantic clustering methods (Cluster):} We include semantic entropy (\textbf{SE}) and discrete semantic entropy (\textbf{DSE}).
\end{itemize}

\subsection{Model Settings}
We use \textbf{LLaMA3.1-8B-Instruct}~\cite{grattafiori_the_llama_3} in our experiments. 
Following \cite{farquhar_detecting_hallucinations_in}, we first set the temperature = $0.1$ and generate an answer as the final answer. Then, we set the temperature to be $1$ and generate 20 answers, which are used for methods that need extra samples. We employ the multinomial sampling as the decoding strategy and set top\_k equal to 50. Due to the varying types of questions and domains, we used the same model to determine the correctness of an answer. The prompts used are in \S~\ref{app:prompts}.

\subsection{Illustrative Results} \label{sec: illustrative_results}

\begin{figure*}[htbp]
\centering
\begin{minipage}[t]{0.49\linewidth}
  \centering
  \includegraphics[width=\textwidth]{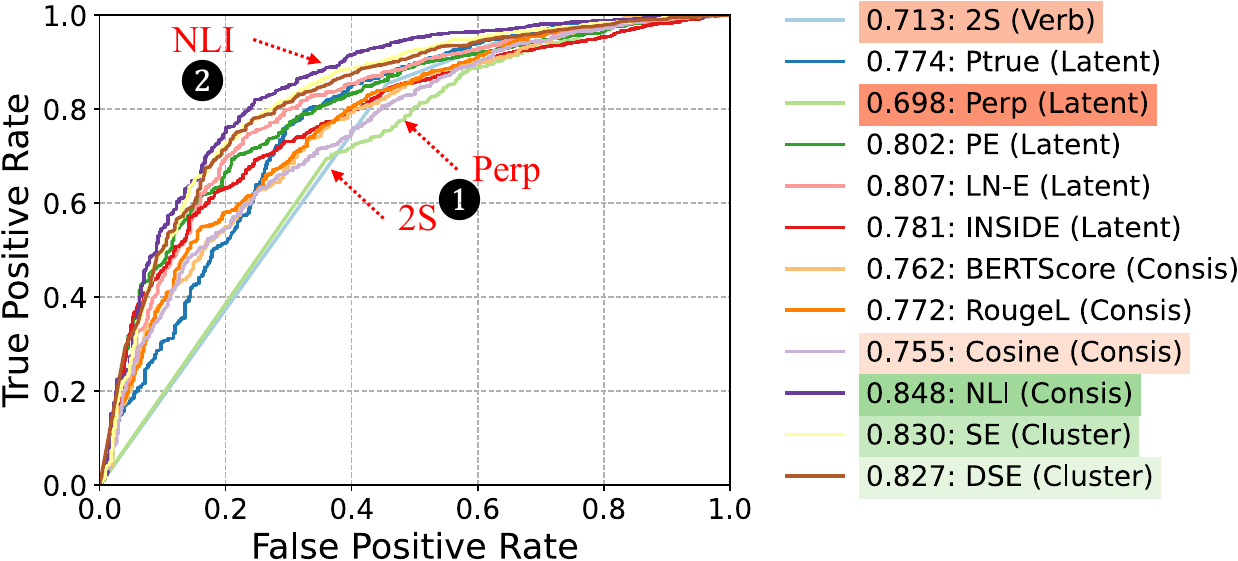}
  \vfill
\end{minipage}\hspace{0.01\linewidth}
\begin{minipage}[t]{0.49\linewidth}
  \centering
  \includegraphics[width=\textwidth]{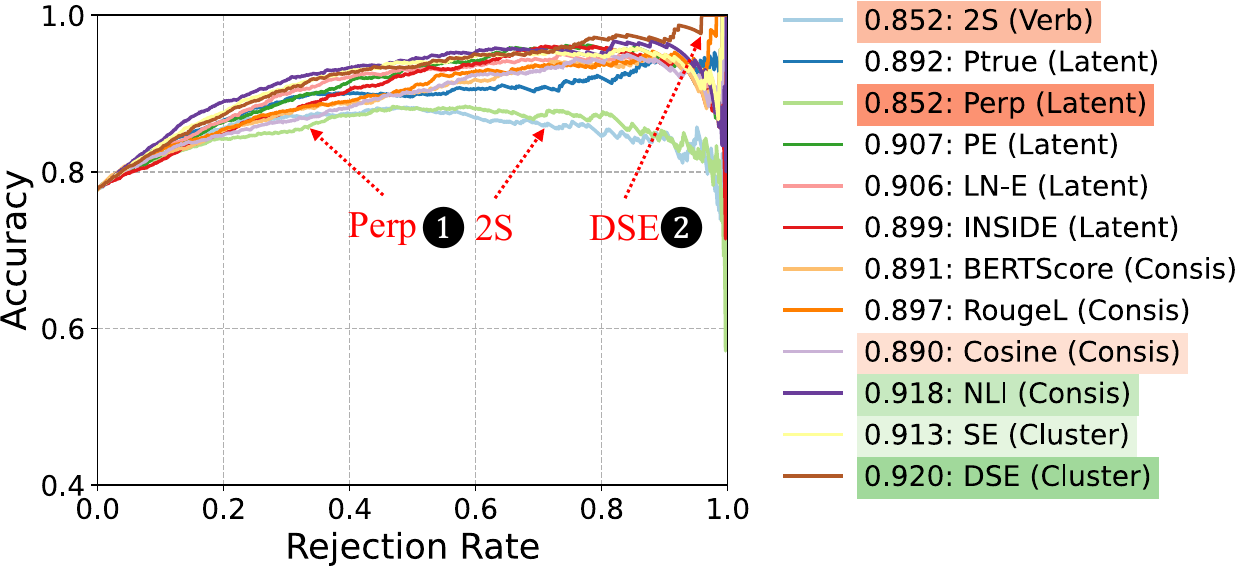}
  \vfill
\end{minipage}
\vspace{-.3in} 
\caption{\textbf{TriviaQA}: ROC (left), ARC (right) curves, and AUROC and AUARC.}
\vspace{-.1in}
\label{fig: trivia_qa}

\vspace{1.5em}

\begin{minipage}[t]{0.49\linewidth}
  \centering
  \includegraphics[width=\textwidth]{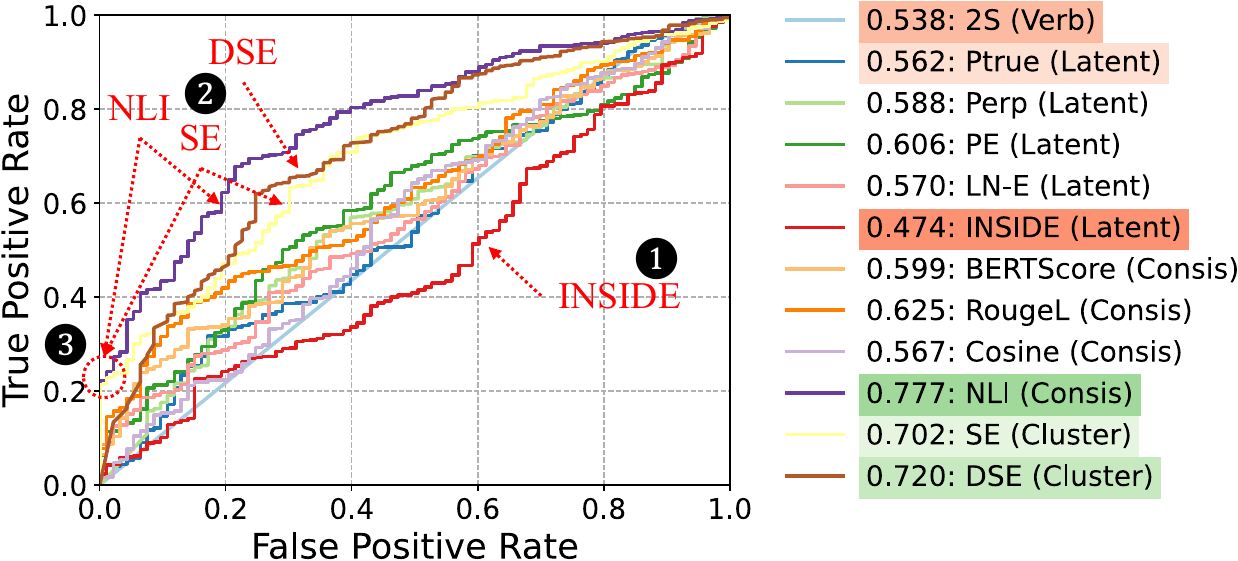}
  \vfill
\end{minipage}\hspace{0.01\linewidth}
\begin{minipage}[t]{0.49\linewidth}
  \centering
  \includegraphics[width=\textwidth]{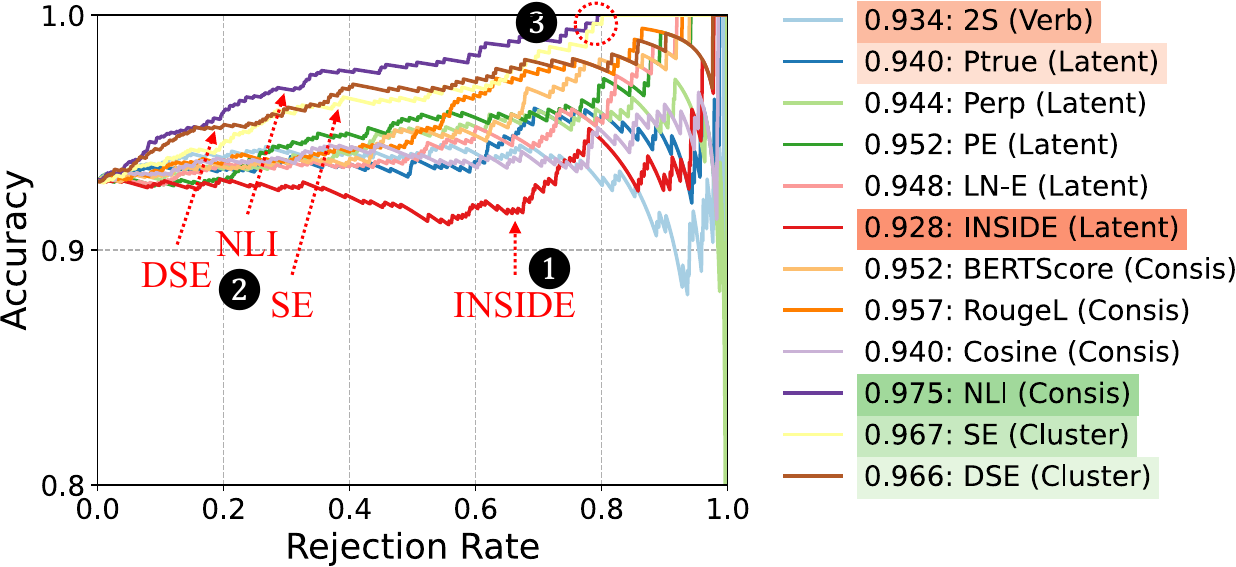}
  \vfill
\end{minipage}
\vspace{-.3in} 
\caption{\textbf{GSM8K}: ROC (left), ARC (right) curves, and AUROC and AUARC.}
\vspace{-.1in}
\label{fig: gsm8k}


\end{figure*}

\begin{figure*}[htbp]
\centering
\begin{minipage}[t]{0.49\linewidth}
  \centering
  \includegraphics[width=\textwidth]{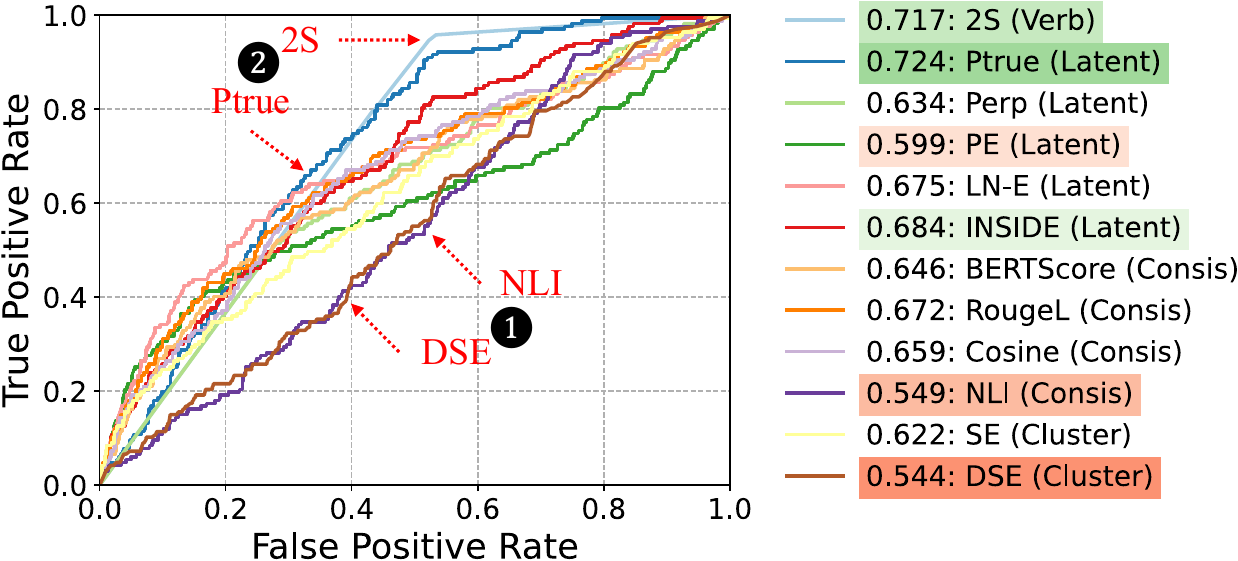}
  \vfill
\end{minipage}\hspace{0.01\linewidth}
\begin{minipage}[t]{0.49\linewidth}
  \centering
  \includegraphics[width=\textwidth]{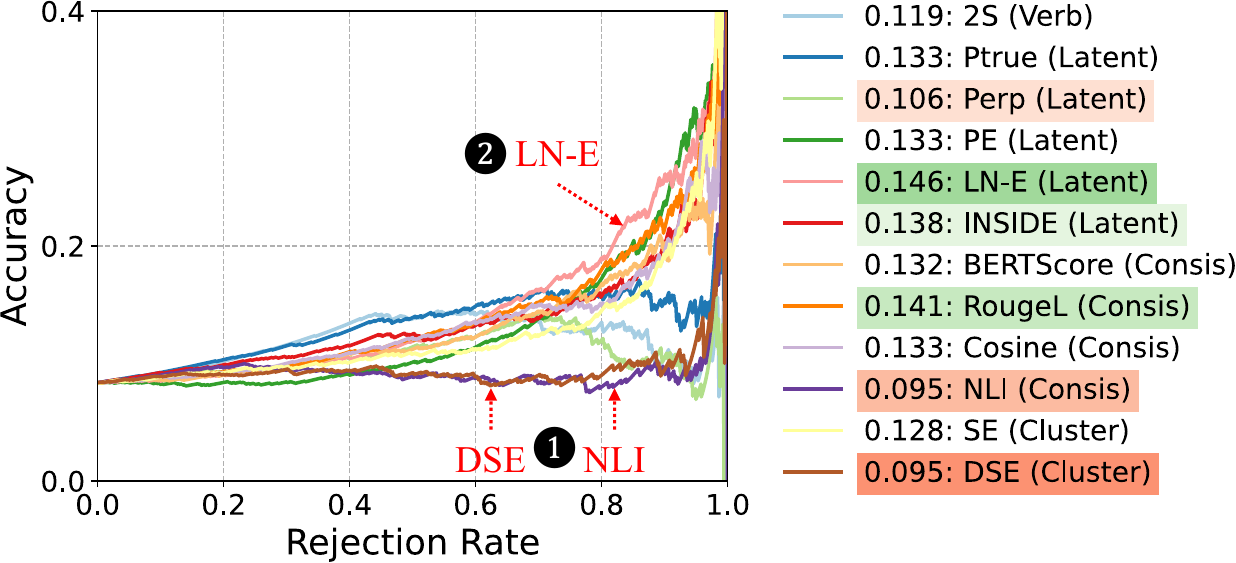}
  \vfill
\end{minipage}
\vspace{-.3in} 
\caption{\textbf{SimpleQA}: ROC (left), ARC (right) curves, and AUROC and AUARC.
\vspace{-.1in}
}
\label{fig: simple_qa}
\end{figure*}

Figures \ref{fig: truthful_qa} - \ref{fig: simple_qa} show the ROC and ARC with the corresponding AUROC and AUARC values in the legend for five different datasets (Details about the datasets are in \S~\ref{app:detailed_dataset}). For the AUROC and AUARC values from the legend, we color-coded the 
\colorbox[HTML]{a1d99b}{``best''}, \colorbox[HTML]{c7e9c0}{``2nd best''}, \colorbox[HTML]{e5f5e0}{``3rd best''}, \colorbox[HTML]{fee0d2}{``3rd worst''}, \colorbox[HTML]{fcbba1}{``2nd worst''}, and \colorbox[HTML]{fc9272}{``worst''}.

\textbf{TruthfulQA} (Figure \ref{fig: truthful_qa}) is a benchmark designed to evaluate the truthfulness of language models in answering questions spanning 38 categories~\cite{lin_truthfulqa}.
The questions in the dataset appear in a multiple-choice form, providing the LLM with clear guidance and ensuring a fixed response format.
Therefore, most of the uncertainty is epistemic uncertainty. \ul{In the ROC curve}, \textbf{Perp} and \textbf{INSIDE} (\blackcircled{1}) demonstrate the lowest performance, close to random guessing. The ROC curve of \textbf{2S} (\blackcircled{2}) starts with the steepest rise, indicating most responses assigned with high confidence are correct. \ul{In the ARC curve}, the worst-performing method (\blackcircled{1}) shows no improvement in accuracy as the rejection rate increases until the rejection rate is high. 
Although \textbf{2S} (\blackcircled{2}) shows a slower initial improvement, it enjoys higher improvements afterward, again demonstrating its high accuracy for high-confidence answers. \textbf{2S} achieves the best performance on this dataset, showing that LLMs can tell their uncertainty, especially when this is mainly epistemic uncertainty.

\textbf{SciQ} (Figure \ref{fig: sciq}) is another multiple-choice Q\&A dataset, with a collection of science-focused questions~\cite{welbl_crowdsourcing_multiple_choice}. \ul{In the ROC curve}, \textbf{Perp} (\blackcircled{1}) performs like random guessing (analogous to \textbf{TruthfulQA}), whereas all other methods achieve significantly better performance, including \textbf{Ptrue} (\blackcircled{2}). Most of the methods (\blackcircled{3}) achieve a very high True Positive Rate (TPR) when the False Positive Rate (FPR) approaches 0.4, indicating they assign most of the low confidence scores to negative samples correctly. \ul{As for AUARC}, most methods exhibit similar performance, as the dataset is considered simple for the LLM, evidenced by a high initial accuracy of about 0.95 (\blackcircled{3}). 
However, the accuracy of \textbf{Perp} (\blackcircled{1}) decreases from the very beginning, resulting in the worst AUARC. In contrast, \textbf{Ptrue} (\blackcircled{2}), another variant of the latent information-based method, gains better accuracy with higher rejection rates. The difference between \textbf{Perp} and \textbf{Ptrue} shows the aggregated predicted probability of tokens is not well-calibrated, but {the probability of answering} the true/false of the entire response is well-calibrated. 

\textbf{TriviaQA} (Figure \ref{fig: trivia_qa}) is a reading comprehension dataset where no context is provided in our settings~\cite{joshi_triviaqa}. As a free-form Q\&A dataset, it allows responses to a question to vary while still expressing the same meaning. Therefore, the aleatoric uncertainty caused by language ambiguity in questions and responses exists. \ul{The ROC curve} reveals that \textbf{2S} and \textbf{Perp} (\blackcircled{1}) demonstrate relatively poor performance. In contrast, \textbf{NLI} (\blackcircled{2}) achieves the highest performance.
\ul{In the ARC curve}, the accuracy of \textbf{2S} and \textbf{Perp} (\blackcircled{1}) deteriorates as the rejection rate increases from 0.5, while \textbf{DSE} (\blackcircled{2}) achieves the highest AUARC score.

\textbf{GSM8K} (Figure \ref{fig: gsm8k}) is comprised of math problems that need reasoning steps to solve~\cite{cobbe_training_verifiers_to}. The responses thus can be more diverse than \textbf{TriviaQA} due to the variability in reasoning steps. Hence, the aleatoric uncertainty is even higher. \ul{The results on AUROC} demonstrate that \textbf{INSIDE} (\blackcircled{1}) performs below random guessing. On the contrary, \textbf{NLI}, \textbf{DSE}, and \textbf{SE} (\blackcircled{2}) maintain more gains on TPR with the increase of FPR. 
A noteworthy observation is that \textbf{NLI} and \textbf{SE} (\blackcircled{3}) achieve positive TPR even when FPR = 0 because they perfectly classify the high-confidence responses.
\ul{In the ARC curve}, this phenomenon is once again reflected that these methods achieve perfect accuracy when considering only the top 20\% high-confidence responses (\blackcircled{3}).
From the point where the rejection rate is 0, better methods exhibit faster rates of improvement (\blackcircled{2}), while the worst one (i.e., \textbf{INSIDE}) has a negative rate (\blackcircled{1}).

Comparing the \textbf{TriviaQA} and \textbf{GSM8K} datasets, \textbf{NLI}, \textbf{SE}, and \textbf{DSE} perform the best on the free-form questions. They all consider the entailment relationship among responses, which can tremendously eliminate the aleatoric uncertainty and thus better estimate epistemic uncertainty. By doing so, they obtain better final results.

\textbf{SimpleQA} (Figure \ref{fig: simple_qa}) is a recent Q\&A dataset that presents significant challenges for state-of-the-art LLM models as of 2024 \cite{wei_measuring_shortform}. Interestingly, \ul{in the ROC curve}, methods (\blackcircled{1}) that traditionally demonstrate superior performance on other datasets exhibit notably poor outcomes here.
\textbf{2S} and \textbf{Ptrue} (\blackcircled{2}) emerge as the top performers, distinguished by their ability to maintain low FPR while TPR approaches 1.
\ul{In the ARC curve}, there is no accuracy improvement for \textbf{NLI} and \textbf{DSE} as the rejection rate increases (\blackcircled{1}). 
Notably, \textbf{LN-E} (\blackcircled{2}) becomes the highest because its accuracy continues to grow after the rejection rate passes 0.8, while others drop. 
Although \textbf{SimpleQA} is still a free-form dataset, \textbf{NLI}, \textbf{SE}, and \textbf{DSE} do not show their superior performance here. It shows they cannot estimate the epistemic uncertainty well if it is too big, postulating whether current benchmarks adequately evaluate LLM uncertainty estimation.

\section{Future Directions} \label{sec: future directions}

\textbf{Uncertainty estimation benchmark.}
We need a dataset specifically designed for uncertainty estimations on LLMs.
Existing datasets are designed to evaluate the capability of LLMs (not their uncertainty). They always have unambiguous questions, resulting in low aleatoric uncertainty. We anticipate three rules for designing this dataset: First, it should incorporate a diverse set of question types, including general Q\&A problems, math problems, translation problems, etc. Second, the questions should have varying difficulty levels, from simple to extremely challenging. 
Finally, the dataset can control the degree of ambiguity for the questions to directly evaluate the uncertainty.

\textbf{Uncertainty estimation method enhancement.}
Uncertainty estimation for long responses remains under-explored. While some papers propose to break long responses into shorter segments and process each part individually \cite{zhang_luq_long_text, farquhar_detecting_hallucinations_in}, they ignore the inter-sentence relationships that are critical for capturing the overall uncertainty of the response. Further, the large vocabulary in long responses challenges the effectiveness of consistency-based and semantic clustering methods.
Current uncertainty estimation methods, predominantly validated on short-answer scenarios, may not adequately address the complexities inherent in longer, multi-step reasoning processes. 

\section{Conclusion} \label{sec: conclusion}
This survey paints a comprehensive landscape for uncertainty estimation methods on LLMs during the inference stage, classifying them into four classes: verbalizing, latent information, consistency-based, and semantic clustering methods. We further enrich our survey with extensive evaluations and promising future directions. 

\section{Limitations}
This survey contains three limitations, mainly due to space constraints. First, we omitted detailed methodological explanations for various methods from the main text. Second, we did not evaluate and report the results of all the introduced methods. 
Finally, we exclude the literature that does not surround the inference stage of LLMs.
We acknowledge these limitations and remain open to academic discussion and collaborative efforts to address them in future work. 

\section*{Acknowledgement}
We thank the anonymous reviewers and the Area Chair for their helpful suggestions and feedback. 
This work was in part supported by the NSF CRII
Award No. 2331536, CAREER Award No. 2326141, and NSF Awards 2212370,
2319880, 2328948, 2411294, and 2417750.

\newpage

\bibliography{custom}

\appendix

\section{Mathematical Formulation of the Methods}
\subsection{Common notations}
We list some common notations in Table~\ref{tab:notation} for mathematical definitions.

\begin{table}[ht]
\centering
\resizebox{\columnwidth}{!}{
\begin{tabular}{l|l}
\hline
\textbf{Notation} & \textbf{Description} \\
\hline
$\bm{f}$ & Large language model \\
\hline
$\bm{x}$ & Input \\
\hline
$\bm{r_i}$ & The $i$-th sampled response \\
\hline
$\bm{r_{*}}$ & The most-voted response from samples \\
\hline
$N$ & Number of tokens in a response \\ 
\hline 
$M$ & Number of sampled responses \\
\hline
$\bm{R_i}$ & The $i$-th response cluster \\
\hline
$K$ & Number of response clusters \\
\hline
$z_i$ & the $i$-th token in a response \\
\hline
$\bm{Z}$ & Vocabulary of the large language model  \\
\hline
$\bm{r_{<i}}$ & All tokens before the $i$-th token \\
\hline
$\bm{p_i}$ & The probability distribution for the $i$-th token \\
\hline
$p_{z_i}$ & The probability for the token $z_i$ \\
\hline
$p$ & The probability of something \\ 
\hline
$a(r_i, r_j)$ & Similarity score between $r_i$ and $r_j$ \\ 
\hline
$U$ & Estimated uncertainty \\
\hline
$C$ & Estimated overall confidence score \\
\hline
$C_i$ & Estimated confidence score for response $\bm{r_i}$ \\
\hline
\end{tabular}
}
\caption{Common notations and descriptions.}
\label{tab:notation}
\end{table}
\subsection{Latent Information Methods} \label{app:formula_latent}

\noindent \textbf{Average over negative logarithm likelihood (Perplexity)}~\cite{manakul_selfcheckgpt, ren_out_of_distribution}:
\begin{equation*}
    U = -\frac{1}{|\bm{r}|}\sum_{i=1}^N \log{p_{z_i}} 
\end{equation*}

\noindent \textbf{Maximum over negative logarithm likelihood}~\cite{manakul_selfcheckgpt}:
\begin{equation*}
    U = \max_i (-\log{p_{z_i}}),\quad i \in [1, N]
\end{equation*}

\noindent \textbf{Ptrue}~\cite{kadavath_language_models_mostly}:
\begin{equation*}
    C = p(z_{true}|\bm{x'}),
\end{equation*}
where $z_{true}$ is the token for ``true'', and $x'$ is the designed prompt to ask LLM to decide whether the answer is true or false.

\noindent \textbf{Predictive entropy}~\cite{kadavath_language_models_mostly}:
\begin{equation*}
    U = -\frac{1}{M}\sum_{j=1}^M \sum_{i=1}^{|\bm{r_j}|}\log p_{z_i} 
\end{equation*}

\noindent \textbf{Length-normalized entropy}~\cite{malinin_uncertainty_estimation_in}:
\begin{equation*}
    U = -\frac{1}{M}\sum_j^M \frac{1}{|\bm{r_j}|} \sum_{i=1}^{|\bm{r_j}|} \log{p_{z_i}}
\end{equation*}

\noindent \textbf{Average over tokens' probability distributions}~\cite{manakul_selfcheckgpt, ren_out_of_distribution}:
\begin{equation*}
    U = - \frac{1}{|\bm{r}|} \sum_{i=1}^N \sum \bm{p_i} \circ \log{\bm{p_i}},
\end{equation*}
where $\circ$ is the element-wise multiplication, and the second $\sum$ means sum over all the elements in a vector.

\noindent \textbf{Maximum over tokens' probability distributions}~\cite{manakul_selfcheckgpt}:
\begin{equation*}
    U = \max_i (-\sum \bm{p_i} \circ \log{\bm{p_i}}),\quad i \in [1, N],
\end{equation*}
where $\circ$ is the element-wise multiplication, and $\sum$ means sum over all the elements in a vector.

\noindent \textbf{INSIDE}~\cite{chen_inside_llms_internal}:
\begin{equation*}
    U = \frac{1}{N} \log \det(\bm{\Sigma} + \alpha \bm{I}) = \frac{1}{N} \sum_{i=1}^N\log(\lambda_i),
\end{equation*}
where $\bm{\Sigma}$ is the covariance matrix, $\alpha$ is a small regularization term, $\bm{I}$ is an identity matrix, and $\lambda_i$ is the $i$-th eigenvalue of the matrix $\bm{\Sigma} + \alpha \bm{I}$. Specifically, 
\begin{equation*}
    \bm{\Sigma} = \bm{V} \cdot \bm{J}_d \cdot \bm{V},\quad \bm{V} = [\bm{v_1}, \bm{v_2}, \cdots, \bm{v_N}],
\end{equation*}
where $\bm{v_i}$ is the representative embedding for $r_i$, $\bm{J}_d = \bm{I}_d - \frac{1}{d}\bm{1}_N\bm{1}_N^T$ represents the centering matrix, and $d$ corresponds to the dimension of the embeddings.

\subsection{Consistency-based Methods} \label{app:formula_consis}
\noindent \textbf{Sampling diversity}~\cite{cole_selectively_answering_ambiguous}:
\begin{equation*}
    C= 1 - \frac{K}{M}
\end{equation*}

\noindent \textbf{Sampling diversity}~\cite{cole_selectively_answering_ambiguous}:
\begin{equation*}
    C = \frac{1}{M} \sum_{i=1}^M \mathds{1}(\bm{r_i}=\bm{r_{*}}),
\end{equation*}
where $\mathds{1}()$ is the indicator function.

\noindent \textbf{First-second-distance-based (FSD) method}~\cite{lyu_calibrating_large_language}:
\begin{equation*}
    C = \frac{1}{M} \sum_{i=1}^M \mathds{1}(\bm{r_i}=\bm{r_{*}}) - \frac{1}{M} \sum_{i=1}^M \mathds{1}(\bm{r_i}=\bm{r_{**}}),
\end{equation*}
where $\mathds{1}()$ is the indicator function, and $\bm{r_{**}}$ denotes the second most-voted answer.

\noindent \textbf{Variation ratio (VR) (when the final response is $\bm{r_{*}}$)}~\cite{huang_look_before_you}:
\begin{equation*}
    U = 1 - \frac{\sum_{i=1}^M \frac{\sum_{j=1, j\neq i}^{j=M}a(\bm{r_i}, \bm{r_j})}{M-1}}{M}
\end{equation*}

\noindent \textbf{Variation ratio (VR) (when the final response is $\bm{r_{M+1}}$)}~\cite{huang_look_before_you}:
\begin{equation*}
    U = 1 - \frac{\sum_{i=1}^M a(\bm{r_i}, \bm{r_{M+1}})}{M}
\end{equation*}

\noindent Based on \textbf{VR} and \textbf{VRO}, using different similarity calculation methods for $a(\cdot, \cdot)$ can achieve different estimated uncertainty. 

\subsection{Semantic Clustering Methods} \label{app:formula_semantic}

\begin{table*}[t]
\begin{tabular}{m{0.11\textwidth}|m{0.05\textwidth}|m{0.56\textwidth}|m{0.17\textwidth}} 
\hline
\textbf{Dataset}    & \textbf{Year}    &\textbf{Question}   & \textbf{Answer}                                          \\ \hline
\textbf{TruthfulQA} & 2021 & \begin{tabular}[c]{@{}l@{}}What are some cities that get more rainfall than Seattle?\\ A) Miami and New Orleans get more rainfall than Seattle.\\ B) No cities get more rainfall than Seattle.\\ C) Seattle gets the most rain.\\ D) It's always raining in Seattle.\end{tabular}         & A) Miami and New Orleans get more rainfall than Seattle. \\ \hline
\textbf{SciQ} & 2017   & \begin{tabular}[c]{@{}l@{}}Who proposed the theory of evolution by natural selection?\\ A) darwin\\ B) Linnaeus\\ C) shaw\\ D) Scopes\end{tabular}                                                                                                                                        & A) darwin                                                \\ \hline
\textbf{TriviaQA} & 2017  & Who was the man behind The Chipmunks?                                                                                                                                                                                                                                                     & David Seville                                            \\ \hline
\textbf{GSM8K} & 2021     & Janet’s ducks lay 16 eggs per day. She eats three for breakfast every morning and bakes muffins for her friends every day with four. She sells the remainder at the farmers' market daily for \$2 per fresh duck egg. How much in dollars does she make every day at the farmers' market? & 18                                                       \\ \hline
\textbf{SimpleQA} & 2024  & Who received the IEEE Frank Rosenblatt Award in 2010?                                                                                                                                                                                                                                     & Michio Sugeno                                            \\ \hline
\end{tabular}
\caption{Samples from each dataset.}
\label{tab:data_sample}
\end{table*}

\noindent \textbf{Semantic entropy}~\cite{kuhn_semantic_uncertainty_linguistic}:
\begin{equation*}
    U = -\sum_{k=1}^K p(\bm{R_k})\log p(\bm{R_k}),
\end{equation*}
where 
\begin{equation*}
    p(\bm{R_k}) = \sum_{\bm{r_j} \in \bm{R_k}} \exp (\frac{1}{|\bm{r_j}|}\sum_{i=1}^{|\bm{r_j}|} \log{p_{z_i}})
\end{equation*}

\noindent \textbf{Discrete semantic entropy}~\cite{farquhar_detecting_hallucinations_in}:
\begin{equation*}
    U = -\sum_{k=1}^K p(\bm{R_k})\log p(\bm{R_k}), 
\end{equation*}
where 
\begin{equation*}
    p(\bm{R_k}) = |\bm{R_k}| / K
\end{equation*}

\section{Detailed Explanation of AUROC and AUARC}
\label{app:detail_au}

\textbf{AUROC}: 
For each response, we consider it as a positive sample (correct) or a negative sample (incorrect) based on whether it matches the ground-truth label. The ROC curve is then created by plotting the true positive rate (TPR) against the false positive rate (FPR). To derive TPRs and FPRs, the accepted confidence threshold is changed to get different Predicted Positives and Negatives (i.e., PP and PN), where a response with confidence higher than the threshold is regarded as PP or PN otherwise.
The AUROC is the area under the ROC curve, measuring the discriminability of confidence scores to distinguish between correct and false responses. 

\textbf{AUARC}: 
Accuracy-Rejection Curve (RAC) is specifically designed for uncertainty estimation, which plots how the accuracy on the accepted samples changes as more low-confidence answers are rejected. The area under it indicates the uncertainty estimation's ability to maintain high accuracy when low-confidence answers are rejected. 

\section{Prompts} \label{app:prompts}

The prompt for Q\&A questions is as follows: 
\begin{center}
\begin{tabular}{p{0.9\linewidth}}
\hline
\textbf{System:}\\
You are a highly knowledgeable assistant. Answer the following question as briefly as possible.\\
\textbf{... (several few-shot examples)}\\
\textbf{User:}\\
\text{[Question]}\\
\hline
\end{tabular}
\end{center}

The prompt for correctness decisions is as follows:
\begin{center}
\begin{tabular}{p{0.9\linewidth}}
\hline
\textbf{User:}\\
We are assessing the quality of answers to the following question: [Question]\\
The expected answer is: [Gt\_answer]\\
The proposed answer is: [Predicted\_answer]\\
Within the context of the question, does the proposed answer mean the same as the expected answer? Respond only with yes or no.\\
Response:\\
\hline
\end{tabular}
\end{center}

\section{Detailed Explanation of Dataset} \label{app:detailed_dataset}

\begin{table*}[]
\resizebox{\textwidth}{!}{
\begin{tabular}{cccccccccccccc}
\hline
\textbf{Dataset}     & \textbf{Metric} & \textbf{2S} & \textbf{Ptrue} & \textbf{Perp} & \textbf{PE} & \textbf{LN-E} & \textbf{INSIDE} & \textbf{BERTScore} & \textbf{RougeL} & \textbf{Cosine} & \textbf{NLI} & \textbf{SE} & \textbf{DSE} \\ \hline
\textbf{AESLC}       & \textbf{AUROC}  & 0.530       & 0.526          & 0.506         & 0.585       & 0.593         & 0.578           & 0.576              & 0.582           & 0.593           & 0.543        & 0.584       & 0.554        \\
\textbf{AESLC}       & \textbf{AUARC}  & 0.383       & 0.387          & 0.358         & 0.435       & 0.435         & 0.430           & 0.421              & 0.426           & 0.439           & 0.406        & 0.426       & 0.403        \\
\textbf{WMT14 De-En} & \textbf{AUROC}  & 0.529       & 0.688          & 0.479         & 0.490       & 0.636         & 0.502           & 0.624              & 0.621           & 0.662           & 0.613        & 0.639       & 0.568        \\
\textbf{WMT14 De-En} & \textbf{AUARC}  & 0.785       & 0.881          & 0.789         & 0.797       & 0.855         & 0.800           & 0.847              & 0.846           & 0.866           & 0.832        & 0.855       & 0.806        \\ \hline
\end{tabular}
}
\caption{Supplementary Results on AESLC and WMT14 De-En.}
\label{tab:supple_results}
\end{table*}

We give a sample for each dataset in Table~\ref{tab:data_sample}.

\noindent \textbf{TruthfulQA}~\cite{lin_truthfulqa} is a benchmark designed to evaluate the truthfulness of language models in generating answers to questions. It consists of 817 questions spanning 38 diverse categories such as health, law, finance, and politics. The dataset is intentionally crafted with questions that humans may answer falsely due to misconceptions or false beliefs

\noindent \textbf{SciQ}~\cite{welbl_crowdsourcing_multiple_choice} is a dataset with 13,7K multiple-choice science questions spanning topics such as biology, chemistry, earth science, and physics. We chose to test our method using its validation set, which contains 1K samples.

\noindent \textbf{TriviaQA}~\cite{joshi_triviaqa} is a large-scale reading comprehension benchmark containing over 650K question-answer-evidence triples, designed to challenge models with complex, compositional questions and diverse evidence sources. In our experimental setup, we do not provide context to the LLM but directly ask it the questions. We selected 2K samples from the validation set for testing.

\noindent \textbf{GSM8K}~\cite{cobbe_training_verifiers_to} is a dataset of 8.5K high-quality linguistically diverse grade school math word problems. Each problem requires 2 to 8 steps to solve, using elementary arithmetic operations ($+, -, \times, \div$). In our experiments, we included reasoning steps in the examples provided in the prompts, and we used their test dataset, which consists of 1.32K samples. 

\noindent \textbf{SimpleQA}~\cite{wei_measuring_shortform} is a benchmark consisting of 4,326 short, fact-seeking questions designed to evaluate the factual accuracy of large language models. It covers a diverse range of topics, including science, politics, art, and so on. The Latest LLMs showed poor accuracy and calibration results on this result. We used 2K samples from the dataset for testing.

\section{Supplementary Results} \label{app:detailed_dataset}
We supplement our experiments by including two additional datasets to evaluate summarization (AESLC~\cite{aeslc}) and machine translation (WMT14 De-En dataset~\cite{wmt14}) tasks. The results are shown in Table~\ref{tab:supple_results}.
The results on the AESLC dataset are consistent with those observed on the SimpleQA dataset, as both exhibit high epistemic uncertainty.
The aleatoric uncertainty for the machine translation task lies between that of the multiple-choice Q\&A and free-form Q\&A datasets. Consequently, we observe comparable performance between consistency-based methods and semantic clustering methods on this task.

\end{document}